\documentclass{article}

\usepackage{PRIMEarxiv}

\usepackage[utf8]{inputenc} 
\usepackage[T1]{fontenc} 

\usepackage{enumitem}
\usepackage{hyperref}       
\usepackage{url}            
\usepackage{booktabs}       
\usepackage{amsfonts}       
\usepackage{nicefrac}       
\usepackage{subfig}   
\usepackage{xcolor}
\usepackage{multirow} 
\usepackage{microtype}      
\usepackage{lipsum}
\usepackage{amsmath} 
\usepackage{fancyhdr}  
\usepackage{array}
\usepackage{pifont}
\usepackage{graphicx}       
\graphicspath{{media/}}     
\newcolumntype{C}{>{\centering\arraybackslash}p{1.2cm}}
\newcolumntype{L}{>{\raggedright\arraybackslash}p{2.8cm}}
\title{Efficient Remote Sensing Change Detection \\with Change State Space Models
}

\author{
  Elman Ghazaei \\
  Faculty of Engineering and Natural Sciences (VPALab),\\
  Sabanci University, T\"{u}rkiye\\
  \texttt{elman.ghazaei@sabanciuniv.edu} \\
   \And
  Erchan Aptoula \\
  Faculty of Engineering and Natural Sciences (VPALab), \\
  Sabanci University, T\"{u}rkiye\\
  \texttt{erchan.aptoula@sabanciuniv.edu} \\
\thanks{This paper has been published in IEEE Geoscience and Remote Sensing Letters (GRSL). Please refer to the published version.}
}

\begin{document}
\maketitle

\begin{abstract}
ConvNets and Vision Transformers (ViTs) have been widely used for change detection, though they exhibit limitations: long-range dependencies are not effectively captured by the former, while the latter are associated with high computational demands. Vision Mamba, based on State Space Models, has been proposed as an alternative, yet has been primarily utilized as a feature extraction backbone. In this work, the Change State Space Model (CSSM) is introduced as a task-specific approach for change detection, designed to focus exclusively on relevant changes between bi-temporal images while filtering out irrelevant information. Through this design, the number of parameters is reduced, computational efficiency is improved, and robustness is enhanced. CSSM is evaluated on three benchmark datasets, where superior performance is achieved compared to ConvNets, ViTs, and Mamba-based models, at a significantly lower computational cost. The code will be made publicly available at https://github.com/Elman295/CSSM upon acceptance. 
\end{abstract}


\section{Introduction}
Change detection (CD) refers to the study of changes on the Earth's surface by analyzing remote sensing images captured from the same region at different time points.
It has a wide range of applications, including urban development planning, land cover change analysis, disaster impact assessment, and ecological monitoring \cite{lu2004change}.

Deep learning has revolutionized CD in remote sensing, with convolutional neural networks (CNN) becoming central for extracting bi-temporal features. In detail, U-Net variants explored strategies like early fusion and mid-level concatenation or subtraction \cite{daudt2018fully}, while CNN-LSTM hybrids modeled temporal dynamics \cite{chen2019change}. Deep supervision \cite{zhang2020deeply} and Siamese networks like SNUNet \cite{fang2021snunet} enhanced feature discrimination using shared weights and dense skip connections. Moreover, attention-based models such as HANet \cite{han2023hanet} and the Change Guiding Network \cite{han2023change} improved focus on relevant regions. Spatiotemporal fusion networks \cite{huang2024spatiotemporal} further boosted performance in complex scenes. Additionally, spiking neural networks (SNN) were utilized in Spiking-UNet \cite{ye2025energy} for SAR-based change detection. However, CNNs and SNNs remain limited in modeling long-range dependencies \cite{chen2024changemamba}.

The emergence of Vision Transformers (ViTs) \cite{dosovitskiy2020image} has provided an effective solution to the limitations of CNNs by leveraging multi-head self-attention to capture long-range dependencies. Building on their success in various computer vision tasks, ViTs have been rapidly adopted for remote sensing CD. A method where CNNs first extract patch-level tokens, which are then processed by transformer encoders and decoders to produce change maps is introduced in \cite{chen2021remote}. A Swin transformer-based model is presented in \cite{zhang2022swinsunet} to capture both local and global features. More recently, the Cross-Temporal Difference Transformer \cite{zhang2023relation} was proposed, which explicitly models cross-temporal relations between bi-temporal images using attention mechanisms, enabling the detection of subtle and meaningful changes.  However, ViTs exhibit quadratic computational complexity, rendering them inefficient for real-time and resource-constrained applications in CD.

Mamba \cite{gu2023mamba}, an evolution of the S4 state space model, introduced input-dependent gating for efficient information retention, achieving strong performance in vision and language tasks with linear computational complexity. Its success has prompted adaptations for remote sensing CD, such as ChangeMamba \cite{chen2024changemamba} that integrates Mamba blocks in encoder-decoder stages but processes bi-temporal images symmetrically, limiting change sensitivity. Other notable examples include Frequency-Enhanced Mamba \cite{xing2025frequency} leveraging frequency cues to capture subtle differences but lacking spatial focus on changed areas. CDMamba \cite{zhang2024cdmamba} on the other hand employs dual streams for temporal modeling yet omits change-guided mechanisms. ConMamba \cite{dong2024conmamba} combines CNNs and Mamba for efficient spatiotemporal learning but lacks explicit change-awareness. Finally, RS-Mamba \cite{zhao2024rs} introduced a low-rank design to reduce overhead, while CW-Mamba \cite{liu2025cwmamba} uses a CNN-Mamba hybrid. However, both still struggle to differentiate changed regions structurally, resulting in suboptimal attention and redundant computation.


In this letter, \textbf{C}hange \textbf{S}tate \textbf{S}pace \textbf{M}odel (CSSM) is proposed specifically for optical remote sensing CD. It enhances the original Mamba by refining its feature extraction to focus exclusively on relevant features associated with changed areas while eliminating irrelevant information. To achieve this, an L1 distance-based approach is employed, incorporating two distinct selection parameters to effectively isolate and extract features corresponding to regions where changes have occurred. By integrating this targeted selection mechanism, CSSM improves detection accuracy while drastically reducing parameters—achieving state-of-the-art performance against ConvNets, ViTs and alternative Mamba based counterparts across multiple benchmark datasets with up to 21.25$\times$ fewer parameters than ChangeMamba models \cite{chen2024changemamba}.



    

\section{Proposed method}

\begin{figure*}[ht]
    \begin{center}
    \includegraphics[width=1.0\textwidth]{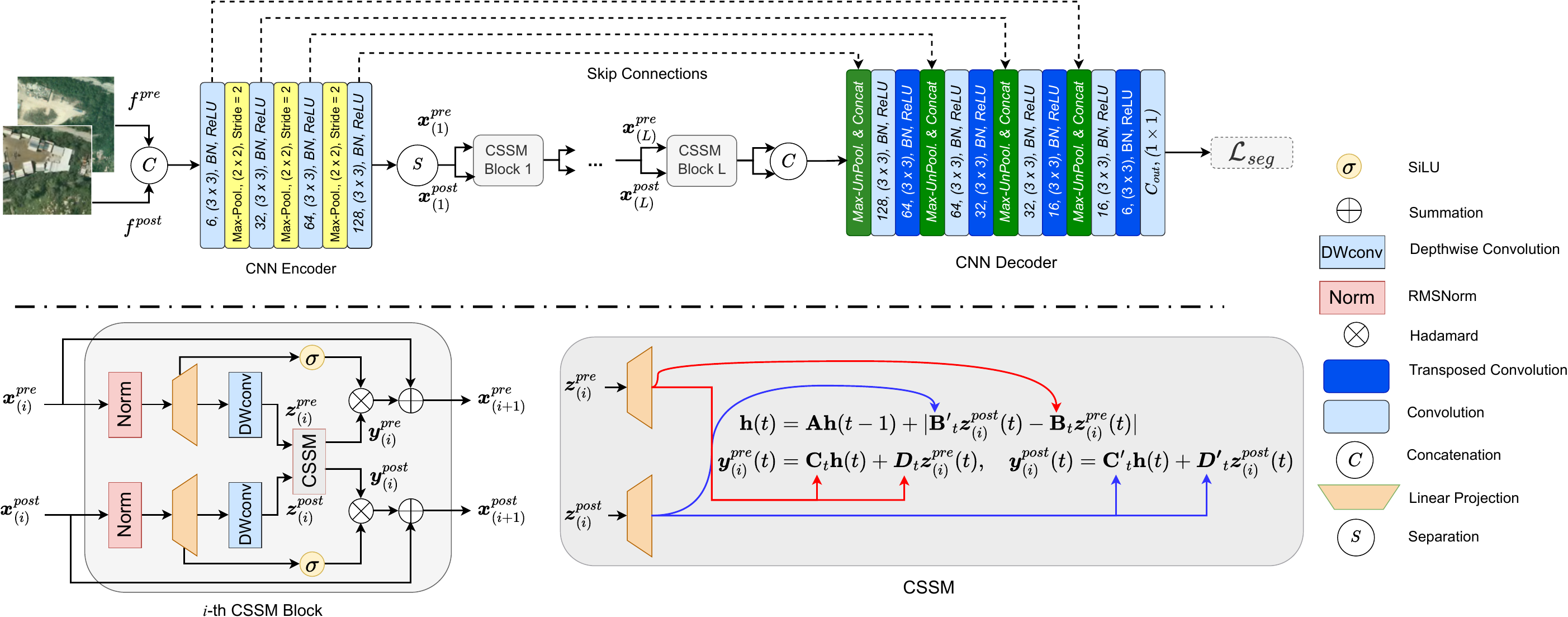}
    \caption{Overview of the proposed framework at the top, and details of the CSSM blocks and CSSM selection mechanism at the bottom half. The CSSM blocks are integrated between a couple of lightweight encoder and decoder, to selectively extract only the most relevant target features.}
    \label{fig:main_fig}
    \end{center}
\end{figure*}

\textbf{Architecture overview:} The proposed architecture (Fig.~\ref{fig:main_fig}) takes pre- ($f^{pre}$) and post-event ($f^{post}$) images as input and consists of three parts: a CNN encoder, CSSM blocks, and a CNN decoder. The features produced by the encoder comprising 4 convolutional blocks, are split into pre- ($\boldsymbol{x}^{pre}$) and post-event ($\boldsymbol{x}^{post}$) representations, which are then processed by sequential CSSM blocks to focus on relevant changed features while filtering out irrelevant information. Finally, the decoder upsamples the features to produce binary change maps.

\textbf{CSSM Block:} each of the sequentially connected CSSM blocks is tasked with progressively enriching the representation of the pre-event and post-event feature maps through the proposed CSSM selection mechanism, in order to emphasize and refine the joint representation of changed content. In detail, each block first receives the normalized $\boldsymbol{x}^{pre}$ and $\boldsymbol{x}^{post}$ and a shared-weight linear layer is applied, followed by a depthwise convolution, producing $\boldsymbol{z}^{pre}$ and $\boldsymbol{z}^{post}$, that are subsequently provided to the CSSM selection mechanism.

On the contrary of the original Mamba selection mechanism \cite{gu2023mamba}, CSSM admits two inputs ($\boldsymbol{z}^{pre}$ and $\boldsymbol{z}^{post}$),  due to the bitemporal nature of CD. The continuous form of CSSM can be expressed as:
\begin{equation}
\begin{aligned}
&\mathbf{h}(t) = \mathbf{A}\mathbf{h}(t-1)
  + \big|\big|\mathbf{B}'_t \boldsymbol{z}^{post}(t)
  - \mathbf{B}_t \boldsymbol{z}^{pre}(t)\big|\big|_{1} \\
&\boldsymbol{y}^{pre}(t) = \mathbf{C}_t \mathbf{h}(t)
  + \mathbf{D}_t \boldsymbol{z}^{pre}(t) \\
&\boldsymbol{y}^{post}(t) = \mathbf{C}'_t \mathbf{h}(t)
  + \mathbf{D}'_t \boldsymbol{z}^{post}(t)
\end{aligned}
\end{equation}
\noindent where \( \mathbf{A} \in \mathbb{R}^{N \times N} \) is the evolution parameter (i.e.~how the state evolves on its own), $N$ denotes the dimensionality of the hidden feature space, $ \mathbf{B},
\mathbf{B'}, \mathbf{D}, \mathbf{D'} \in \mathbb{R}^{N \times 1},$ $ \mathbf{C}, \mathbf{C'} \in \mathbb{R}^{1 \times N}$ are the projection parameters, and $||\cdot||_{1}$ stands for the $L_{1}$ norm. Moreover, the input projection was modified by incorporating $||\mathbf{B'_t} \boldsymbol{z}^{post}(t) - \mathbf{B}_t \boldsymbol{z}^{pre}(t)||_{1}$, thus presumably enabling the detection of changed regions, with $\mathbf{B'}_t \boldsymbol{z}^{post}(t)$ and $\mathbf{B}_t \boldsymbol{z}^{pre}(t)$ aiming to extract relevant features from $\boldsymbol{z}^{post}$ and $\boldsymbol{z}^{pre}$ respectively. 
Furthermore, the output projection was modified by incorporating $\mathbf{C}_t \boldsymbol{h}(t) + \mathbf{D}_t \boldsymbol{z}^{pre}(t)$ and $\mathbf{C'}_t \boldsymbol{h}(t) + \mathbf{D'}_t \boldsymbol{z}^{post}(t)$ to integrate relevant features from $\boldsymbol{z}^{pre}$ and $\boldsymbol{z}^{post}$ into the outputs $\boldsymbol{y}^{pre}$ and $\boldsymbol{y}^{post}$. This modification presumably enables residual effects to directly propagate information from the input to the output, allowing for a deeper model architecture and enhancing the overall representational capacity.

After Zero-Order Hold discretization \cite{gu2023mamba}, the discrete CSSM selection mechanism can be expressed as:
\begin{equation}
\begin{aligned}
&\mathbf{h}_t = \mathbf{\overline{A}}\mathbf{h}_{t-1}
  + \big|\big|\mathbf{\overline{B}'}_t \boldsymbol{z}^{post}_t
  - \mathbf{\overline{B}}_t \boldsymbol{z}^{pre}_t\big|\big|_{1}, \\
&\boldsymbol{y}^{pre}_t = \mathbf{C}_t \mathbf{h}_t
  + \mathbf{D}_t \boldsymbol{z}^{pre}_t, \\
&\boldsymbol{y}^{post}_t = \mathbf{C}'_t \mathbf{h}_t
  + \mathbf{D}'_t \boldsymbol{z}^{post}_t
\end{aligned}
\end{equation}
\noindent After the final CSSM block, its output is concatenated and passed through the CNN decoder that produces the final change map.




\textbf{Loss Function:} The cross-entropy function ($\mathcal{L}_{CE}$) was utilized to optimize the model parameters, while the Lovász-Softmax loss ($\mathcal{L}_{lov}$)  was employed to address the issue of imbalance in terms of changed and unchanged pixels counts as per \cite{chen2024changemamba}. The total loss function of the proposed model is:
\begin{equation}
\mathcal{L}_{seg} = \mathcal{L}_{CE} + \mathcal{L}_{lov}
\label{eq:lossseg}
\end{equation}

\section{Experiments}
\label{sec.exp}
\textbf{Datasets:} 1) SYSU-CD \cite{shi2021deeply} is a large-scale dataset containing 20,000 pairs of high-resolution aerial images (0.5 m/pixel) from Hong Kong, capturing diverse urban and coastal changes between 2007 and 2014. It is split into training, validation, and test sets in a 6:2:2 ratio using $256 \times 256$ pixel patches. 2) LEVIR-CD+ \cite{chen2020spatial} possesses 985 pairs of $1024 \times 1024$ pixel high-resolution (0.5 m/pixel) images, featuring complex urban and suburban land-use changes. It is divided into training, validation, and test sets, following the split of \cite{chen2024changemamba}. 3) WHU-CD \cite{ji2018fully} consists of high-resolution (0.3 m/pixel) aerial image pairs, covering a 20.5 $km^2$ area and highlighting significant urban expansion and land-use changes. It follows the official training and testing split defined in \cite{ji2018fully}. 



\begin{table*}[t]
    \begin{center}        
    \caption{Results of CD experiments with the highest values highlighted in \textcolor{red}{red} and the second highest in \textcolor{blue}{blue}. A ``-'' is used for cases where a method has neither a reported performance with a dataset nor an available implementation.}    
    \label{tableNumericalResults}
    \scriptsize{
    \begin{tabular}{lccc|ccc|ccc|ccc}
        \toprule
        & \multicolumn{3}{c}{\textbf{LEVIR-CD +}} & \multicolumn{3}{c}{\textbf{WHU-CD}} & \multicolumn{3}{c}{\textbf{SYSU-CD}} & \textbf{Params (M)} & \textbf{GFLOPs} &

        \multicolumn{1}{c}{\textbf{Inference speed (ms)}}\\
        \midrule
        \multicolumn{12}{c}{\textbf{ConvNet based approaches}}\\
        \midrule
        \textbf{Method}   & \textbf{OA} & \textbf{F1} & \textbf{IoU}   & \textbf{OA} & \textbf{F1} & \textbf{IoU}  & \textbf{OA} & \textbf{F1} & \textbf{IoU}& &&on WHU-CD  \\
        \midrule
        FC-EF \cite{daudt2018fully}   & 97.54 & 70.42 & 54.34   & 98.87 & 84.89 & 73.74   & 88.69 & 75.81 & 61.04 &1.35&14.13 &34.3 \\
        FC-Siam-Diff \cite{daudt2018fully}    & 98.26 & 77.57 & 63.36  & 99.13 & 87.67 & 78.04   & 88.65 & 75.79 & 61.01 &1.35&18.66 & 34.2\\
        FC-Siam-Conc \cite{daudt2018fully}   & 98.24 & 78.44 & 64.53    &98.94 & 85.83 & 75.18  & 88.05 & 75.18 & 60.23 & 1.54&21.07&35.1 \\
        SiamCRNN \cite{chen2019change}  & 98.67 & 83.20 & 71.23   & 99.19 & 89.10 & 80.34  & 90.77 & 80.44 & 67.28 & 63.44 & 224.30 &89.5 \\
        DSIFN \cite{zhang2020deeply}   & 98.70 & 84.07 & 72.52    & 99.31 & 89.91 & 81.67  & 89.59 & 78.80 & 65.02 & 35.73 &329.03 & 82.1\\
        SNUNet \cite{fang2021snunet}   & 97.83& 74.70 & 59.62   & 99.10& 87.70 & 78.09  & 87.49 & 73.14 & 57.66 &10.21 & 176.36 & 64.3\\
        HANet \cite{han2023hanet}   & 98.22 & 77.56 & 63.34   & 99.16 & 88.16 & 78.82    & 89.52 & 77.41 & 63.14 & 2.61&70.68 & 53.0\\
        CGNet  \cite{han2023change}  & 98.63 & 83.68& 71.94    & 99.48 & 92.59& 86.21    & 91.19 & 79.92 & 66.55 & 33.68&329.58 & 60.9\\
        SEIFNet \cite{huang2024spatiotemporal}   & 98.66 & 83.32& 71.41   & 99.36 & 91.29& 83.98  & 89.86 & 78.45 & 64.54 & 39.08 & 167.75 &76.0 \\
        \midrule
        \multicolumn{12}{c}{\textbf{Vision transformer based approaches}}\\
        \midrule
        ChangeFormer \cite{bandara2022transformer}  & 97.60 & 72.71 & 57.12    & 98.83 & 83.66 & 71.91 & 89.67 & 76.76 & 62.29 & 41.03&811.15 & 74.1  \\
        BIT \cite{chen2021remote}  & 98.60 & 82.53 & 70.26   & 99.27 & 90.04 & 81.88  & 90.76 & 79.41 & 65.84 & 33.27& 380.62 & 77.0\\
        TransUNetCD  \cite{li2022transunetcd} & 98.66 & 83.63 & 71.86   & 99.09 & 87.79 & 78.44  & 90.88 & 80.09 & 66.79 & 28.37 & 244.54 &47.3\\
        SwinSUNet \cite{zhang2022swinsunet}   & 98.92 & 85.60 & 74.82   & 99.50 & 93.04 & 87.00  & 91.51 & 81.58 & 68.89 &39.28  &43.50 & 42.5\\
        CTDFormer \cite{zhang2023relation}   & 98.40 & 80.30 & 67.09    & 99.20 & 88.67 & 79.65  & 90.06 & 78.08 & 64.04 & 3.85 & 303.77 & 60.1\\
        \midrule
        \multicolumn{12}{c}{\textbf{Mamba based approaches}}\\
        \midrule
        ChangeMamba \cite{chen2024changemamba}  & \textcolor{blue}{\textbf{99.06}} & 88.39 & 79.20   & \textcolor{blue}{\textbf{99.58}} &94.19 & 89.02    & 92.30 & 83.11 & 71.10 &84.70 & 179.32 & 51.3\\

        ConMamba \cite{dong2024conmamba}   & - & - &-    & 99.39 &93.01  & 86.89   & - & - &- &34.18&85.43 & - \\
        CDMamba \cite{zhang2024cdmamba}    & 98.65 & 83.01 & 70.95   &99.51  & 93.76 & 88.26   & 90.38 & 79.07 &70.33&\textcolor{blue}{\textbf{10.99}}&259.49 &  46.3 \\
        
        RS-Mamba \cite{zhao2024rs} &98.42  & 80.91 & 67.95  & 99.44 & 92.79 & 86.55  & 89.74 & 77.91 & 69.90 &27.9 &50.20 & \textcolor{blue}{\textbf{39.9}}\\

        CWMamba \cite{liu2025cwmamba}   & 98.98 & 87.21 &77.32   & -& -& -   & \textcolor{red}{\textbf{92.96}} & \textcolor{red}{\textbf{84.33}} &\textcolor{blue}{\textbf{72.90}} &373.62 &\textcolor{blue}{\textbf{37.32}} & -  \\

        FE-Mamba \cite{xing2025frequency}   & \textcolor{red}{\textbf{99.12}} & \textcolor{blue}{\textbf{91.16}} &\textcolor{blue}{\textbf{83.24}}  & \textcolor{red}{\textbf{99.62}} & \textcolor{red}{\textbf{95.19}} &\textcolor{blue}{\textbf{90.23}}    & \textcolor{blue}{\textbf{92.80}} & \textcolor{blue}{\textbf{83.97}} & 72.87 & 51.33 & 136.90 & 55.5\\
       \textbf{ CSSM (Ours)}   & 98.69 & \textcolor{red}{\textbf{92.39}} &\textcolor{red}{\textbf{86.63}} & 99.12&\textcolor{blue}{\textbf{94.98}} &\textcolor{red}{\textbf{90.80}}& 88.65 & 83.66 &\textcolor{red}{\textbf{72.95}} & \textcolor{red}{\textbf{4.34}}& \textcolor{red}{\textbf{5.10}} &\textcolor{red}{\textbf{16.4}} \\
        \bottomrule
    \end{tabular}}
    \end{center}
    \begin{center}
    \end{center}
\end{table*}

\textbf{Training:}
All models were trained using the Adam optimizer with an initial learning rate of 0.001, decayed by StepLR every 10 epochs. Training was conducted for 100 epochs using an NVIDIA RTX 4090 GPU.

\subsection{Results}

\textbf{Quantitative Analysis:}
According to the results of Table~\ref{tableNumericalResults}, Mamba-based models consistently outperform CNN and Transformer-based approaches. CSSM, in particular, shows strong generalization across datasets, surpassing previous Mamba variants. Moreover, while other models excel in specific metrics or datasets, CSSM achieves a more balanced and robust performance, highlighting its effectiveness in diverse CD scenarios.

\begin{figure*}[t]
  \begin{center}
  \subfloat[Pre-event] {\label{fig.example1}\includegraphics[width=0.091\textwidth]{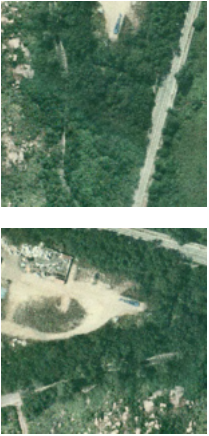}}\,
  \subfloat[Post-event]{\label{fig.example1}\includegraphics[width=0.091\textwidth]{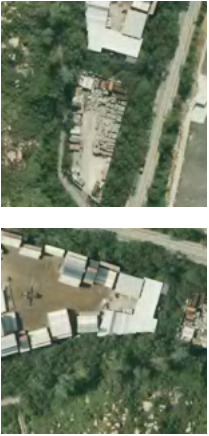}}\,
  \subfloat[GT]{\label{fig.example1}\includegraphics[width=0.091\textwidth]{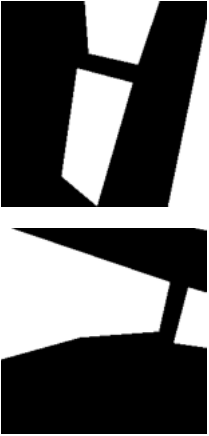}}\,
  \subfloat[FC-Conc]{\label{fig.example1}\includegraphics[width=0.091\textwidth]{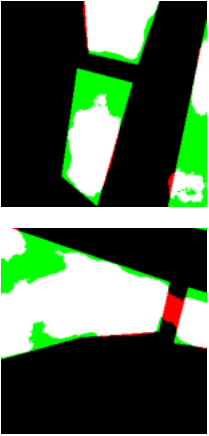}}\,
  \subfloat[SNUNet]{\label{fig.example1}\includegraphics[width=0.091\textwidth]{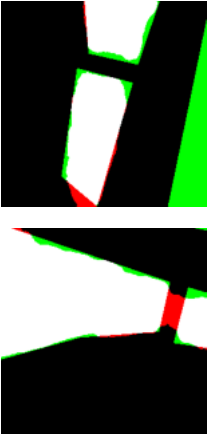}}\,
  \subfloat[Ch.Former]{\label{fig.example1}\includegraphics[width=0.091\textwidth]{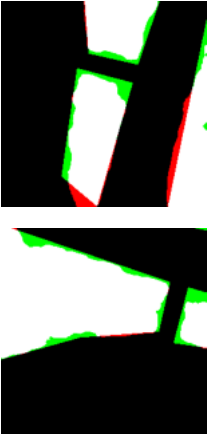}}\,
      \subfloat[BIT]{\label{fig.example1}\includegraphics[width=0.091\textwidth]{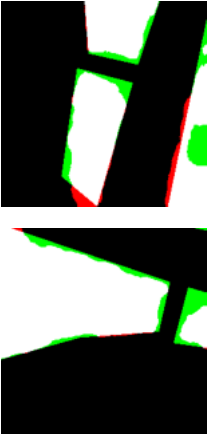}}\,
    \subfloat[Ch.Mamba]{\label{fig.example1}\includegraphics[width=0.091\textwidth]{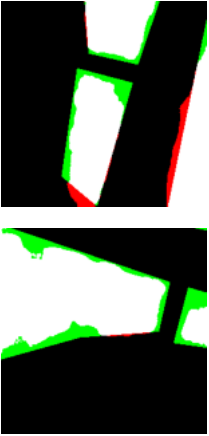}}\,
    \subfloat[CSSM]{\label{fig.example1}\includegraphics[width=0.091\textwidth]{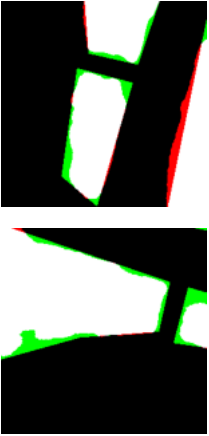}}\,
    \subfloat[CAM]{\label{fig.example1}\includegraphics[width=0.091\textwidth]{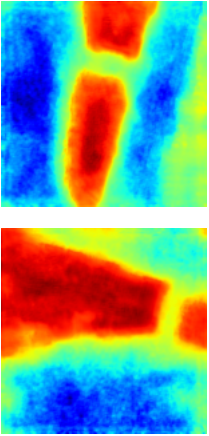}}
  \caption{Qualitative CD results on the SYSU-CD dataset. 
\setlength{\fboxsep}{0pt}\fbox{\textcolor{red}{\rule[0.0ex]{2em}{2ex}}} False Positives, 
\setlength{\fboxsep}{0pt}\fbox{\textcolor{green}{\rule[0.0ex]{2em}{2ex}}} False Negatives, 
\setlength{\fboxsep}{0pt}\fbox{\textcolor{white}{\rule[0.0ex]{2em}{2ex}}} True Positives.}
  \label{fig:examples-sysu}
\end{center}
\end{figure*}

\textbf{Qualitative Analysis:}
Fig.~\ref{fig:examples-sysu} illustrates the estimated pixel-level CD maps for SYSU-CD, in addition to Grad-CAM visualizations that illustrate the model’s ability to focus on relevant regions, confirming its strong feature extraction capabilities and robustness in complex scenarios. These visualizations highlight that the network consistently attends to areas of actual changes.

\textbf{Computational Cost:}
Table~\ref{tableNumericalResults} summarizes the parameter counts, FLOPs and inference speed (for a single image pair on the WHU-CD dataset) of all tested approaches. While reported methods exhibit a wide variability in terms of model size and computational load, CSSM stands out for its efficiency, offering competitive accuracy with significantly fewer parameters and lower resource requirements. Thanks to the effectiveness of the proposed CSSM blocks, even an extremely (computationally) modest encoder-decoder pair enables the proposed architecture to surpass state-of-the-art counterparts at a fraction of the cost; thus rendering it particularly suitable for resource-constrained settings. 



\begin{table}[h]
\begin{center}     
 \caption{Effect of CSSM block count ($L$) on performance.}
       \label{tab:ablation}
    \begin{tabular}{l l c c c c c}
        \toprule
        & \textbf{Method} & \textbf{Rec} & \textbf{Pre} & \textbf{OA} & \textbf{F1} & \textbf{IoU}  \\
        \midrule
        & CNN instead of CSSM&87.34&82.67&98.34&84.94&74.19 \\
        & 0 CSSM Blocks &85.12&83.21&98.79&84.15&72.56 \\
        & 1 CSSM block & 90.05& 91.77& 98.32& 90.89 &84.36  \\
        & 2 CSSM blocks &89.87 &91.93& 98.71&90.61 &84.56 \\
        & 3 CSSM blocks &90.79 &91.84& 98.63&91.54 &84.96  \\
        & 4 CSSM blocks &\textcolor{blue}{\textbf{91.89}} &93.01& \textcolor{red}{\textbf{99.25}}&\textcolor{blue}{\textbf{94.81}} &\textcolor{blue}{\textbf{90.23}}  \\

         & 5 CSSM blocks & \textcolor{red}{\textbf{94.23}}&\textcolor{red}{\textbf{95.75}}& \textcolor{blue}{\textbf{99.12}}&\textcolor{red}{\textbf{94.98}} &\textcolor{red}{\textbf{90.80}} \\

         & 6 CSSM blocks & 90.10&\textcolor{blue}{\textbf{93.81}}&98.96&92.08&85.61  \\

         & 7 CSSM blocks & 91.25&93.86&98.97&92.54&86.34  \\
         
         & 8 CSSM blocks & 90.79&92.50&98.31&92.31&85.52  \\
        \bottomrule      
    \end{tabular}
 \end{center}
\end{table}

\subsection{Ablation Experiments}

\textbf{Number of CSSM blocks:}
Table~\ref{tab:ablation} presents the results of ablation experiments conducted on the WHU-CD dataset, evaluating the impact of different numbers of CSSM layers, as well as the replacement of CSSM by a convnet (128 filters, kernel size 1, stride 1).
The results indicate that performance improves as the number of layers increases up to 5. This setup offers a strong balance across all evaluation metrics. The 4-layer version also performs competitively, particularly in terms of overall accuracy. However, adding more layers beyond this point does not lead to further improvements and may increase model complexity unnecessarily. 

\textbf{Effect of Distance Metric on Performance:}
Alternative distance functions, including L2, Chebyshev Distance and cosine similarity, were also explored (Table~\ref{tab:distance_metrics}). L1 was observed to outperform all, as it is presumably more robust to outliers, and better preserves sparse and high-dimensional feature differences.

\begin{table}[t]
\begin{center}
    
\caption{Effect of metric on CSSM performance (WHU-CD dataset).}
\label{tab:distance_metrics}
\begin{tabular}{lccc}
\toprule
\textbf{Metric} & \textbf{OA} & \textbf{F1} & \textbf{IoU } \\
\midrule
L1 Distance     & \textcolor{blue}{\textbf{99.12}}&\textcolor{red}{\textbf{94.98 }} &   \textcolor{red}{\textbf{90.80 }}     \\
L2 Distance        &  \textcolor{red}{\textbf{99.14}} & \textcolor{blue}{\textbf{94.86}} & \textcolor{blue}{\textbf{90.20}}                 \\
Chebyshev Distance    & 94.12  &  87.63     & 85.20  \\
Cosine Similarity       & 98.56 &   92.14    & 89.05  \\
\bottomrule
\end{tabular}
\end{center}

\end{table}

\begin{figure}[h]
  \begin{center}
  \subfloat[Gaussian Blur]{\label{fig.example1}\includegraphics[width=0.44\textwidth]{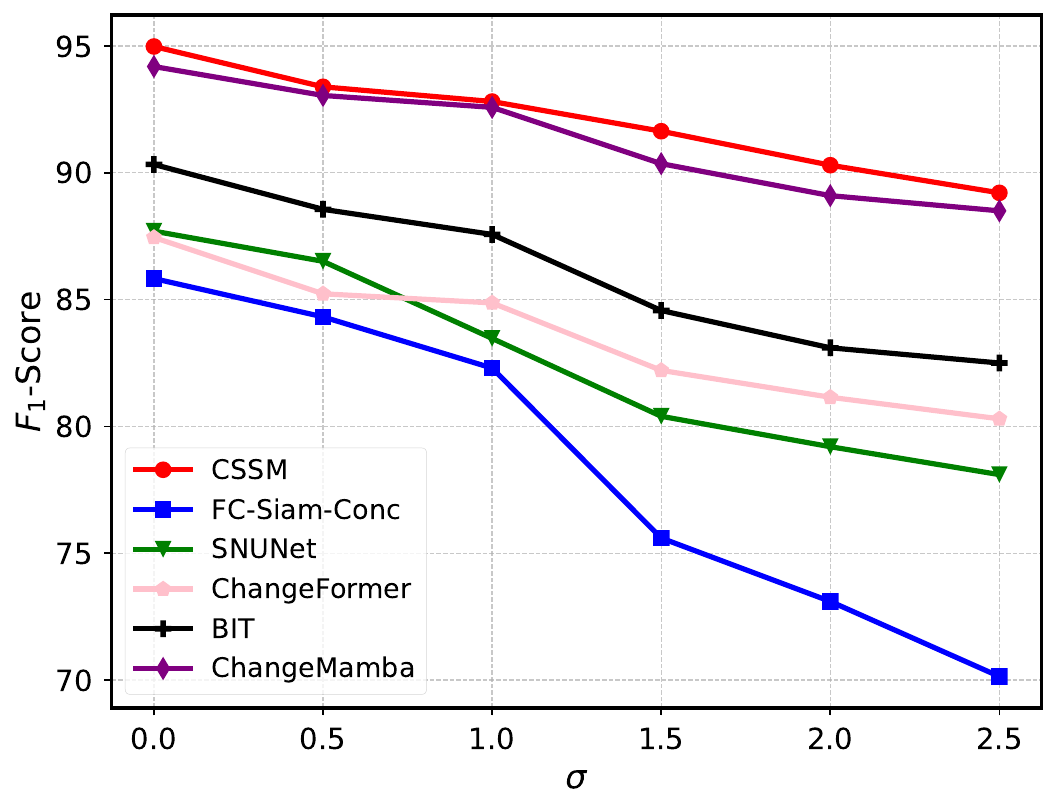}}\,
  \subfloat[Gaussian Noise]{\label{fig.example2}\includegraphics[width=0.44\textwidth]{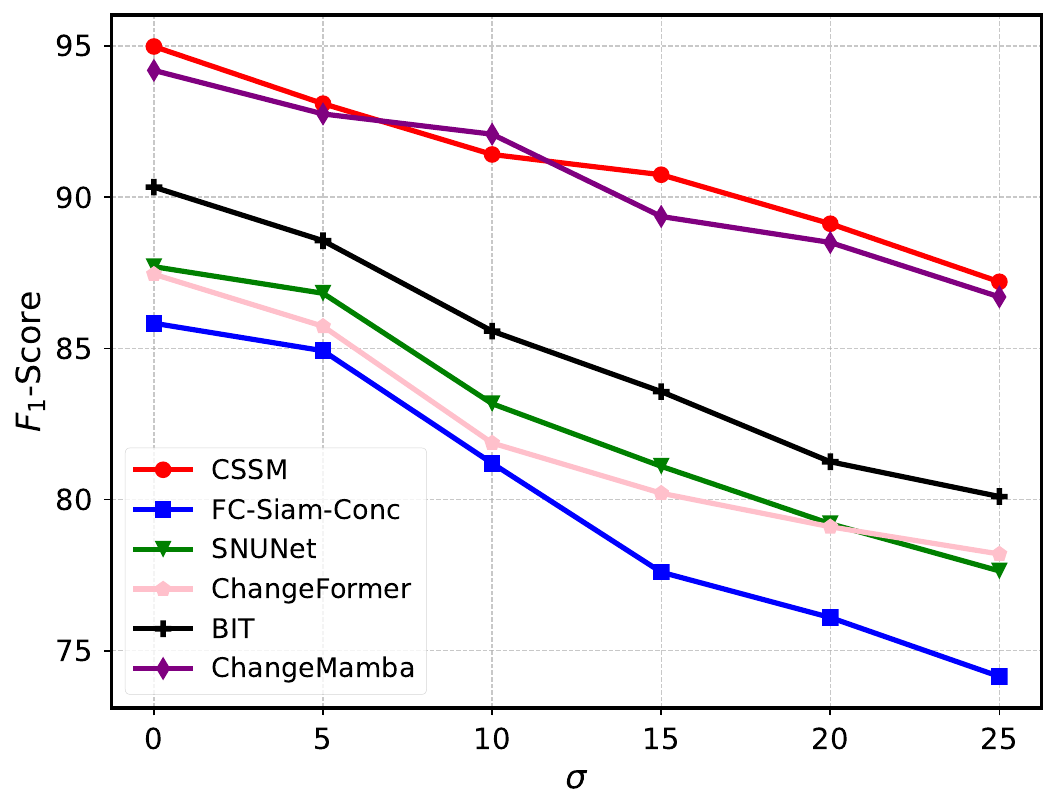}}\,
  \caption{Robustness assessment against degraded WHU-CD data.}   
  \label{fig:abbb}
\end{center}
\end{figure}

\textbf{Robustness Against Degraded Input Data:}

Gaussian blur and additive Gaussian noise of various levels were applied to the WHU-CD dataset images during inference to simulate common real-world imaging artifacts (Fig.~\ref{fig:abbb}). CNN-based models showed sensitivity to these perturbations, with F1-scores declining significantly, probably because their features are highly dependent on precise pixel-level details. In contrast, transformer-based architectures exhibited greater resilience, with only moderate performance reductions, likely due to their ability to capture long-range contextual information and dependencies. CSSM and ChangeMamba, on the other hand, consistently provided the highest robustness. CSSM, in particular, maintained strong and stable performance across both types of degradation, reflecting its capacity to focus on semantically relevant features and effectively disregard localized noise or blur. These findings underscore the practical applicability of CSSM in real-world remote sensing scenarios, where image quality can often be compromised due to adverse weather, sensor limitations, or acquisition conditions.

\textbf{Robustness against Seasonal and Weather Variations:} To assess the robustness of CSSM against diverse seasonal variations and adverse weather conditions, the CDD dataset \cite{lebedev2018change} consisting of bi-temporal optical high-resolution Google Earth images was chosen. The experimental setup used in \cite{xing2025frequency} was reproduced and the results (Table \ref{tab:real_world}) indicate that CSSM shows superior performance.

\textbf{Dealing with multiple modalities:} To assess the performance of CSSM with multi-modal data, the OSCD dataset \cite{daudt2018urban} was employed, which comprises co-registered optical and synthetic aperture radar (SAR) image pairs captured before and after change events. The experimental setup used in \cite{zhang2025cdprompt} was reproduced and even though CSSM was not developed with multi-modality in mind (the optical and SAR images were concatenated in the case of CSSM), the results (Table \ref{tab:oscd}) show that it still leads to a performance comparable to the state-of-the-art.


\begin{table}[h]
\begin{center}
\caption{Performance of the proposed CSSM on the CDD dataset, with the highest values highlighted in \textcolor{red}{red} and the second highest in \textcolor{blue}{blue} against various baseline approaches \cite{xing2025frequency}.}
\label{tab:real_world}
\begin{tabular}{lcccc}
\toprule
\textbf{Method} & \textbf{Pre} &\textbf{Rec}& \textbf{F1} & \textbf{OA} \\
\midrule
    SNUNet &92.26 & 95.48& 95.87 & 98.98  \\
     HFA-Net & 76.52 &71.34   &73.84& 93.76             \\
      SGSLN & \textcolor{blue}{\textbf{97.16}}&  \textcolor{red}{\textbf{96.88}}   & \textcolor{blue}{\textbf{97.02}} & \textcolor{blue}{\textbf{99.27}} \\
 STNet  & 95.79&  93.45   & 94.60 &98.68  \\
 BIT     &91.77 & 91.70    &91.74  & 97.96 \\
ChangeFormer  & 92.46& 90.04    &91.24  &97.86  \\
ACABFNet& 95.75& 94.79    & 95.27 & 98.84 \\
ChangeViT& 97.02&  96.27   & 94.64 & 99.17 \\
ChangeMamba &95.98 & \textcolor{blue}{\textbf{96.41}}    &96.19  &99.06  \\
CDMamba &95.82 & 94.50    &95.15  &98.81  \\
RS-Mamba & 95.92&  96.33   & 96.12 & 99.04 \\
\textbf{CSSM (Ours)}&\textcolor{red}{\textbf{98.14}} &   96.35  &  \textcolor{red}{\textbf{97.25}}& \textcolor{red}{\textbf{99.30}} \\
\bottomrule
\end{tabular}
\end{center}
\end{table}

\begin{table}[h]
\begin{center}
\caption{Performance of the proposed CSSM on the OSCD dataset, with the highest values highlighted in \textcolor{red}{red} and the second highest in \textcolor{blue}{blue} against various baseline approaches \cite{zhang2025cdprompt}.}
\label{tab:oscd}
\begin{tabular}{lccc}
\toprule
\textbf{Method} & \textbf{Pre} &\textbf{Rec}& \textbf{F1}  \\
\midrule
    FC-EF &10 & \textcolor{red}{\textbf{60}}& 17.1  \\
    FC-Siam-Diff &13.2&21.3&16.8 \\
    FC-Siam-Conc &12.7&26.6&17.2 \\
    Fres-UNet &20.3&21.7&21.1 \\
    MMCD &17.3&20.8&18.9\\
    DS-UNet &19.6&21.7&20.6\\
    SmaDA &\textcolor{red}{\textbf{26.4}}&17.3&20.9\\
    Siamese-UNet &15.4&29.1&20.1\\
    CDPrompt &\textcolor{blue}{\textbf{25.8}}&37.6&\textcolor{blue}{\textbf{30.6}}\\
    \textbf{CSSM (Ours)} &25.5&\textcolor{blue}{\textbf{42.1}}&\textcolor{red}{\textbf{31.8}}\\
\bottomrule
\end{tabular}
\end{center}
\end{table}

\section{Conclusion}
\label{sec:con}
This letter has introduced CSSM as a novel architecture for remote sensing CD. By leveraging the efficiency of state space models, it was designed to effectively capture relevant changes between bi-temporal images while filtering out irrelevant information. This formulation allowed for the elimination of redundant network parameters, thereby reducing computational complexity without compromising detection accuracy. Extensive experiments were conducted on three benchmark datasets, where CSSM exhibited superior performance compared to CNN-based, Transformer-based, and Mamba-based approaches. The results demonstrated that high accuracy could be achieved while ensuring scalability, making CSSM a promising solution for large-scale remote sensing CD application. Future work will focus on the integration of domain generalization techniques to enhance generalization across diverse remote sensing datasets.

\bibliographystyle{unsrt}  
\bibliography{ref}

\end{document}